\documentclass{article}

\usepackage{microtype}
\usepackage{graphicx}
\usepackage{subcaption}
\usepackage{booktabs} % for professional tables

\usepackage{hyperref}

\usepackage[preprint]{icml2026}

\usepackage{amsmath}
\usepackage{amssymb}
\usepackage{mathtools}
\usepackage{amsthm}
\usepackage{multirow}
\usepackage{color, colortbl}
\usepackage{algorithmic}
\usepackage{algorithm}
\renewcommand{\algorithmiccomment}[1]{\hfill $\triangleright$ #1}
\usepackage{tabularx}

\usepackage[capitalize,noabbrev]{cleveref}

\theoremstyle{plain}

\theoremstyle{definition}

\theoremstyle{remark}

\usepackage[textsize=tiny]{todonotes}

\icmltitlerunning{DiscoForcing: A Unified Framework for Real-Time Audio-Driven Character Control with Diffusion Forcing}

\begin{document}

\twocolumn[
  % \icmltitle{DiscoForcing: Streaming Audio-Driven Diffusion for Real-Time Character Control}

\icmltitle{DiscoForcing: A Unified Framework for Real-Time Audio-Driven\\ Character Control with Diffusion Forcing}

% Your Music Dance Twin, from Avatar to Humanoid 

% \icmltitle{DiscoForcing: A Unified Framework for Real-Time Music-to-Dance with Diffusion Forcing}

% \icmltitle{DiscoForcing: A Real-Time Music-to-Dance Motion Synthesis Framework with Diffusion Forcing}

  % It is OKAY to include author information, even for blind submissions: the
  % style file will automatically remove it for you unless you've provided
  % the [accepted] option to the icml2026 package.

  % List of affiliations: The first argument should be a (short) identifier you
  % will use later to specify author affiliations Academic affiliations
  % should list Department, University, City, Region, Country Industry
  % affiliations should list Company, City, Region, Country

  % You can specify symbols, otherwise they are numbered in order. Ideally, you
  % should not use this facility. Affiliations will be numbered in order of
  % appearance and this is the preferred way.

    \icmlsetsymbol{equal}{*}
    \icmlsetsymbol{Corresponding author}{$\dagger$}
    
    \begin{icmlauthorlist}
    \icmlauthor{Kaiyang Ji}{shanghaitech,comp,equal}
    \icmlauthor{Bingsheng Qian}{shanghaitech,comp,equal}
    \icmlauthor{Binghuan Wu}{shanghaitech}
    \icmlauthor{Kangyi Chen}{shanghaitech}
    \icmlauthor{Ye Shi}{shanghaitech,comp}
    \icmlauthor{Jingya Wang}{shanghaitech,Corresponding author}
    \end{icmlauthorlist}
    
    \icmlaffiliation{shanghaitech}{ShanghaiTech University}
    \icmlaffiliation{comp}{InstAdapt}
    \icmlcorrespondingauthor{Jingya Wang}{wangjingya@shanghaitech.edu.cn}
  % You may provide any keywords that you find helpful for describing your
  % paper; these are used to populate the "keywords" metadata in the PDF but
  % will not be shown in the document
  \icmlkeywords{Machine Learning, ICML}

  \vskip 0.3in
]

% this must go after the closing bracket ] following \twocolumn[ ...

% This command actually creates the footnote in the first column listing the
% affiliations and the copyright notice. The command takes one argument, which
% is text to display at the start of the footnote. The \icmlEqualContribution
% command is standard text for equal contribution. Remove it (just {}) if you
% do not need this facility.

% Use ONE of the following lines. DO NOT remove the command.
% If you have no special notice, KEEP empty braces:
\printAffiliationsAndNotice{\icmlEqualContribution}   % no special notice (required even if empty)
% Or, if applicable, use the standard equal contribution text:
% \printAffiliationsAndNotice{\icmlEqualContribution}

\begin{abstract}
We study real-time audio-responsive character control as a deployment-faithful problem: strictly causal, bounded-latency streaming that must generate coherent full-body motion at interactive frame rates while the audio condition can change abruptly (tempo shifts, drops, or user edits). Prior music-to-motion systems are largely optimized for offline generation with global context, and degrade in streaming rollouts where conditioning history becomes stale or unreliable. We introduce DiscoForcing, a streaming audio-driven diffusion framework that combines a causal music encoder that captures rhythmic structure and phase dynamics with a diffusion-forcing sequence model trained under heterogeneous noise levels across the temporal horizon. Building on this, we design a hybrid temporal schedule and a history-guided streaming sampler to explicitly trade off responsiveness against long-horizon consistency under non-stationary audio. Implemented in an end-to-end real-time interactive system with online avatar playback and humanoid deployment workflows, DiscoForcing delivers more stable long-horizon rollouts and sharper audio–motion alignment than prior baselines under matched causality and latency constraints while maintaining real-time throughput. Project Page: \url{https://discoforcing.github.io/}
\end{abstract}

%%%%%%%%%%%%%%%%%% Introduction %%%%%%%%%%%%%%%%%%%%%%%
\begin{figure}[t]
    \centering
    \includegraphics[width=\linewidth]{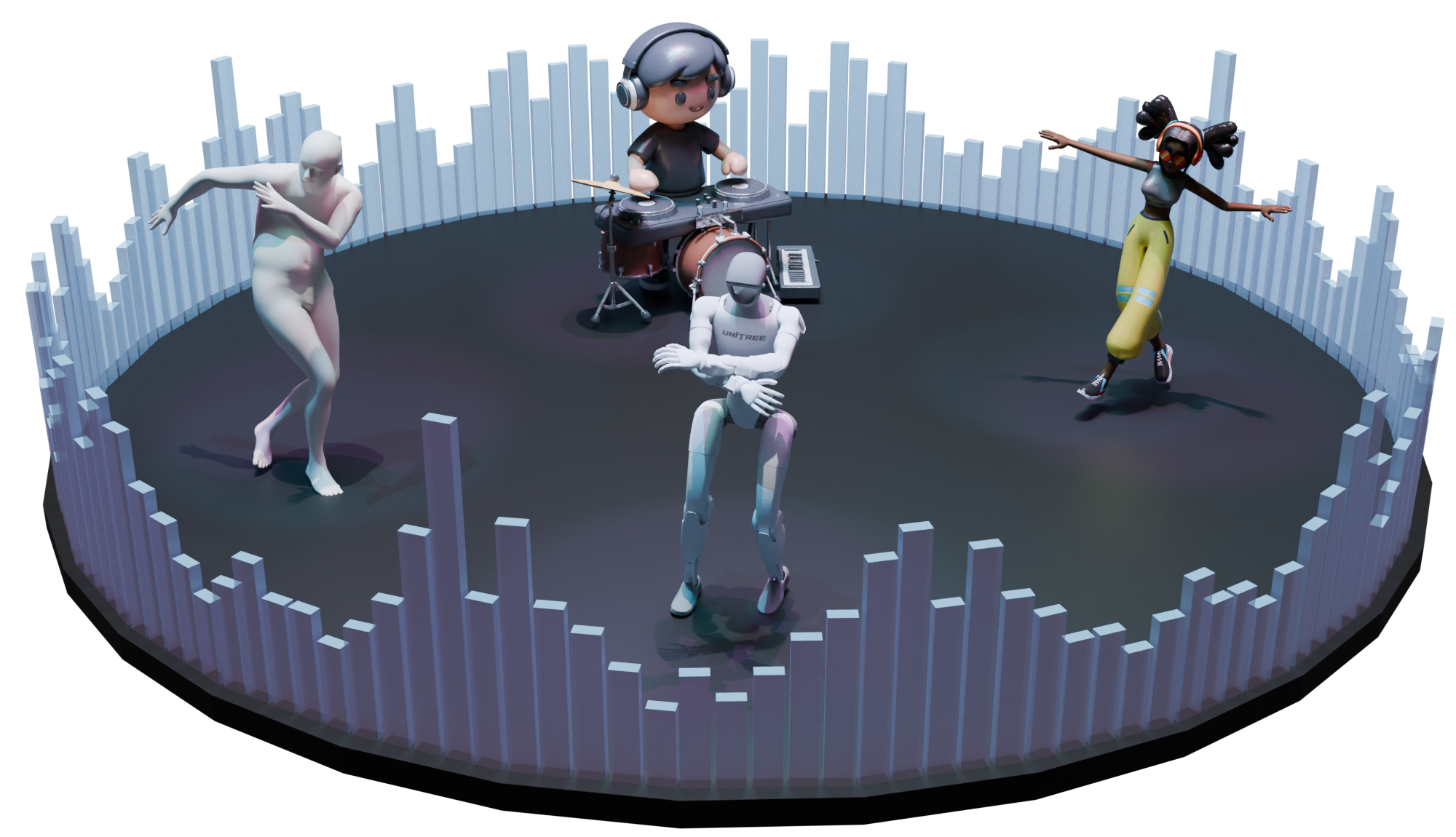}
    \caption{We introduce \textbf{DiscoForcing}, a real-time, audio-responsive character control system. Given online streaming audio inputs, DiscoForcing causally synthesizes continuous full-body motions in real time. The generated motion supports two deployment settings: (i) \emph{avatar interactive control} for responsive animation and visualization, and (ii) \emph{physics-based humanoid platform} by converting the predicted motion into executable humanoid joint commands.}
    \label{fig:teaser}
\end{figure}
\section{Introduction}
\label{sec:intro}

\begin{table*}[t]
\centering
\caption{Comparative analysis of DiscoForcing versus existing audio-driven human motion synthesis approaches.}
\resizebox{1.0\textwidth}{!}{
\begin{tabular}{l|ccccc}
\toprule
Method & online system & long-horizon & music transition & physical platform & user-interactive \\
\midrule
Bailando~\cite{siyao2022bailando} &  & \checkmark &  &  &  \\
EDGE~\cite{tseng2023edge}    &  & \checkmark & \checkmark &  &  \\
Lodge~\cite{li2024lodge}     &  & \checkmark &  &  &  \\
MEGADance~\cite{yang2025megadance} &  &  & \checkmark &  &  \\
FlowerDance~\cite{yang2025flowerdance}&  &  &  &  & \checkmark \\
\midrule
DiscoForcing & \checkmark & \checkmark & \checkmark & \checkmark & \checkmark \\
\bottomrule
\end{tabular}}
\label{tab:task-com}
\end{table*}

Real-time audio-responsive character control is a core capability for interactive embodied applications, spanning virtual reality avatars, animation-driven simulation, and physical humanoid deployment. In these settings, a system must convert a live music stream into continuous full-body motion that is simultaneously (i) \emph{responsive} to newly arriving audio cues and user edits, (ii) \emph{temporally coherent} over long rollouts, and (iii) \emph{computationally feasible} at interactive frame rates under a strict per-frame budget. Beyond entertainment, this problem provides a concrete and deployment-faithful testbed for studying streaming generative policies under causality and latency constraints, which aligns with the broader goal of making generative sequence models usable in real interactive loops.

Despite rapid progress in music-to-motion generation, most prior methods are developed and validated in an offline regime. They commonly assume access to future audio context~\cite{alexanderson2023listen, li2021ai}, rely on long temporal windows~\cite{siyao2022bailando}, or perform non-causal inference that can revisit and refine past decisions~\cite{li2024lodge}. While these choices can improve offline metrics on fixed clips, they do not translate cleanly to streaming deployment. When executed online with zero lookahead, models often exhibit delayed reactions~\cite{xiao2025motionstreamer}, weakened beat synchronization~\cite{tseng2023edge}, and accumulating artifacts in long-horizon rollouts~\cite{holden2017phase}. Importantly, this mismatch is structural rather than incidental: streaming execution changes the learning and inference problem because the model must act causally on incomplete evidence, commit to outputs without revision, and satisfy a hard real-time compute budget~\cite{zhaodartcontrol}.

Streaming audio-responsive control is challenging for three intertwined reasons. First, causality and bounded latency remove the anticipatory cues that offline systems exploit for beat alignment and phrase-level structure, making instantaneous synchronization substantially harder~\cite{siyao2022bailando, barquero2024seamless}. Second, music streams are non-stationary: tempo shifts, drops, style switches, and user-driven edits can induce abrupt conditioning changes. Over-trusting motion history improves smoothness but slows reaction, whereas aggressively overriding history increases responsiveness but risks discontinuities and jitter~\cite{fan2025align, huang2024beat}. Third, long-horizon streaming amplifies error accumulation. Any small prediction error contaminates the future conditioning history, inducing a train--test distribution shift that is exacerbated when the audio condition itself evolves over time~\cite{xiao2025motionstreamer, holden2017phase}.
\begin{figure*}[t]
    \centering
    \includegraphics[width=\linewidth]{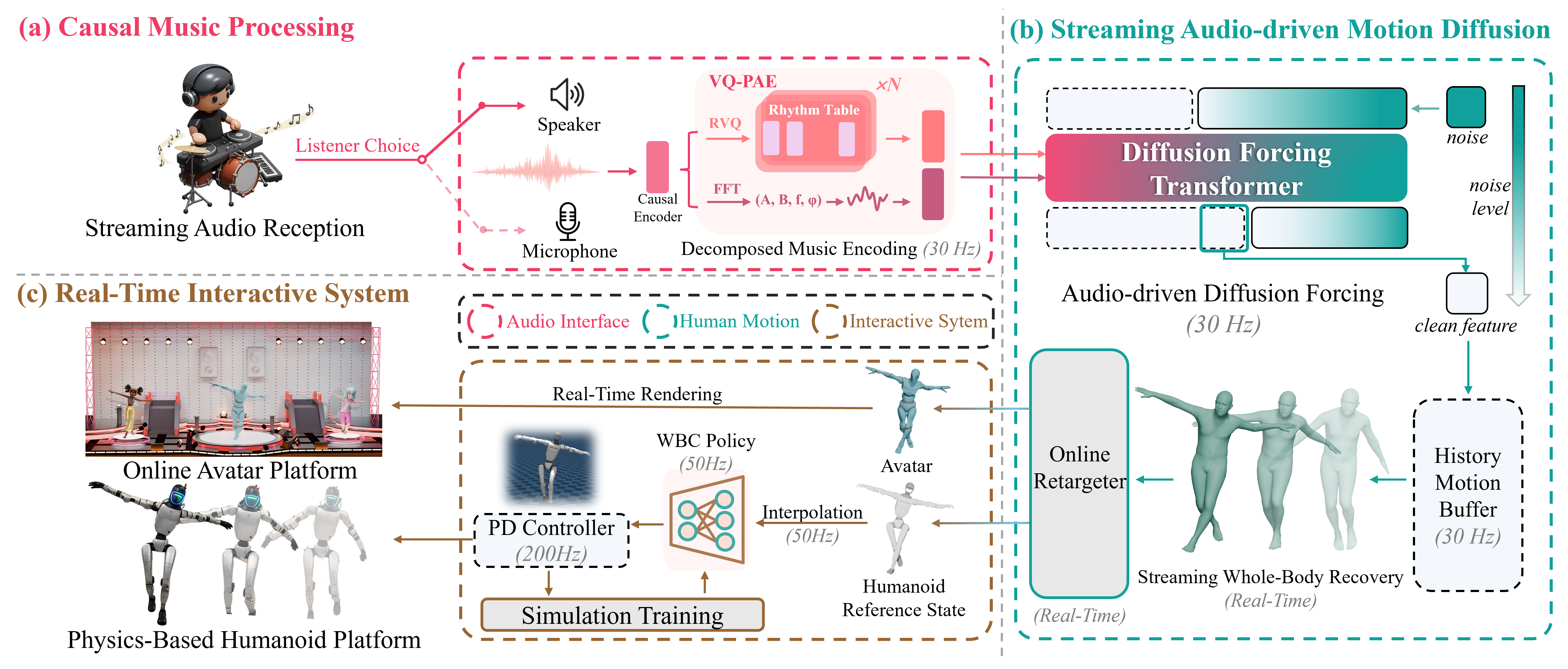}
    \caption{\textbf{System Pipeline.} DiscoForcing encodes live audio into a \emph{causal} music feature (30\,Hz) and generates continuous full-body motion via a diffusion-forcing transformer conditioned on the feature and a history buffer (30\,Hz). The resulting motion is delivered to (i) an online avatar platform for retargeting and interactive Unity visualization, and (ii) a humanoid deployment stack that performs IK/interpolation and executes whole-body control with low-level PD tracking.}
    \label{fig:pipeline}
\end{figure*}

A central limitation of the current literature is the lack of a \textbf{deployment-faithful end-to-end system} for audio-responsive character control. Most prior work is presented as an offline generation model evaluated on fixed clips, rather than a real-time interactive stack that users can directly experience. In practical use, however, the model must run online: it must process streaming audio, produce motion continuously in real time, and commit to outputs without retroactive revision~\cite{dai2024motionlcm}. Without an integrated system that couples perception/encoding, streaming generation, and online execution, it remains unclear how existing models behave when placed in an actual interactive loop, and what design choices are necessary to deliver stable and responsive user-facing control.

In this work, we study real-time audio-responsive character control under explicit streaming constraints. We formulate the task as bounded-latency streaming motion generation: at each time step, the model receives a causal music representation extracted from the live audio stream and a finite motion history buffer, and then predicts the next short motion chunk at interactive rate. This formulation mirrors practical interactive systems and enables principled, apples-to-apples evaluation under matched constraints.

Building on this formulation, we propose \textbf{DiscoForcing}, a streaming audio-driven diffusion framework that addresses the stability--responsiveness trade-off in non-stationary rollouts. DiscoForcing couples (i) a causal music encoder that extracts rhythmic structure and phase dynamics from a sliding audio window with (ii) a diffusion-forcing sequence model trained with heterogeneous noise across time, improving robustness to imperfect or stale autoregressive histories. 
During inference, we use a streaming sampler with hybrid temporal scheduling, where temporal guidance combines a stable history-based prediction with an audio-conditioned prediction that preserves recent context while discounting distant history, thereby reducing lock-in to stale autoregressive states and improving responsiveness to incoming audio.

To close the evaluation gap, we also build an end-to-end real-time online verification system that runs the full loop under strict causality and low-latency constraints. The system enables real-time, audio-driven full-body motion generation from streaming audio inputs and supports two deployment settings: (i) a user-facing avatar platform for responsive animation and interaction, and (ii) a physics-based humanoid pipeline that executes the generated motion as joint-level control commands. This setup enforces matched causality and latency budgets and supports both interactive demos and embodied execution.

% It causally encodes live audio, synthesizes whole-body motion in real time, and uses a unified communication stack to drive two deployment front-ends: (i) a user-facing avatar platform with online skeleton retargeting, and (ii) a physics-based humanoid pipeline that converts motion into joint commands via temporally consistent retargeting and interpolation, followed by real-time tracking control. This setup enforces matched causality and latency budgets and supports both interactive demos and embodied execution.

Under streaming settings, DiscoForcing improves both offline metrics and online rollout behavior compared with strong autoregressive and diffusion baselines. In particular, it yields better responsiveness to abrupt music changes without sacrificing continuity, while maintaining real-time throughput under strict per-frame computation budgets. These results indicate that diffusion-forcing models, when paired with deployment-aware conditioning management and streaming evaluation, offer a practical path toward robust audio-responsive character control in interactive applications, as shown in Table~\ref{tab:task-com}.

Our contributions are threefold:
\begin{itemize}
    \item We present \textbf{DiscoForcing}, to our knowledge, the first end-to-end real-time audio-responsive character control framework that integrates causal audio encoding with streaming dance generation.
    \item We propose a latent diffusion-forcing model with decomposed music conditioning on rhythmic and periodic features, and introduce a temporal-guided strategy to balance fast reaction with long-horizon stability.
    \item We build a reproducible communication-based interactive system that enables real-time user interaction for both online avatar playback and a physics-based humanoid platform, supporting practical demonstrations and cross-platform validation within a unified runtime.
    % Our system bridges the virtual and physical worlds through a unified communication-based platform. It enables seamless, real-time interaction across both digital avatars and physical humanoid robots, supporting cross-platform validation and practical demonstrations.
    
\end{itemize}

\section{Related Works}
\label{sec:related}
\paragraph{Audio-Driven Human Motion Synthesis.}
Human motion synthesis has rapidly advanced with deep generative models that support rich conditioning signals, from action labels~\cite{guo2020action2motion, petrovich2021action}, text descriptions~\cite{tevethuman, chen2023executing, zhang2023generating, jiang2023motiongpt, wangscaling, liu2025commanding} and various interactions~\cite{ji2025towards, tang2024unified, deng2026humanobject} to audio. Audio-driven generation introduces a distinctive synchronization challenge: the model must align kinematic events to fine-grained rhythm while preserving coherent, natural full-body dynamics over long horizons. Early systems such as FACT~\cite{li2021ai} and Bailando~\cite{siyao2022bailando} use autoregressive transformers, but often accumulate errors in long rollouts, degrading beat timing and motion fidelity. Diffusion-based methods improve coverage of complex choreography and stylistic diversity~\cite{alexanderson2023listen, tseng2023edge, li2023finedance}, with further advances in structure and controllability via coarse-to-fine sampling (Lodge~\cite{li2024lodge}), bidirectional context (BADM~\cite{zhang2024bidirectional}), and genre-aware experts (MEGADance~\cite{yang2025megadance}). However, most approaches remain offline and non-causal, relying on future audio or full-track access for global consistency, which limits streaming use where the system must react causally to live audio and user edits under tight latency budgets.

\paragraph{Real-Time Interactive Character Control.}
Interactive character control is a core problem in animation and physics-based simulation, enabling users to steer virtual agents in a responsive and physically plausible way. Early approaches such as DeepPhase~\cite{starke2022deepphase} and PFNN~\cite{holden2017phase} rely on low-dimensional controls (e.g., heading direction, speed, simple style flags) to drive motion generation, yielding smooth locomotion but limited expressivity. Later methods like PADL~\cite{juravsky2022padl} and AnySkill~\cite{cui2024anyskill} use natural language as the control interface, formulating character control as language-conditioned motion generation. Building on this, CAMDM~\cite{chen2024taming}, DART~\cite{zhaodartcontrol} and FloodDiffusion~\cite{cai2025flooddiffusion} employ autoregressive diffusion models for text-driven motion synthesis, producing long-horizon motions with richer multimodality and temporal coherence. In parallel, CLoSD~\cite{tevetclosd} explicitly closes the loop between motion generation and simulation to enforce physical feasibility, while recent work such as Diffuse-CLoC~\cite{huang2025diffuse} and UniPhys~\cite{wu2025uniphys} jointly models state and action distributions to more tightly couple generative priors with control dynamics. However, audio-driven real-time interactive character control remains largely unexplored: music is highly non-stationary and diverse in rhythm and style, and existing systems rarely support a user-facing, real-time interaction and validation platform for evaluating control quality from the end-user’s perspective.

\paragraph{Diffusion Models for Sequence Modeling.}
Diffusion probabilistic models~\cite{sohl2015deep, ho2020denoising} invert a progressive noising process and have become dominant in image generation~\cite{dhariwal2021diffusion, saharia2022photorealistic, rombach2022high}, motivating adaptations to sequential domains. A common approach is full-sequence denoising, jointly refining all timesteps~\cite{ho2022video, janner2022planning, song2023loss, gupta2024photorealistic}. While effective for fixed-length synthesis, such non-causal formulations are ill-suited to variable-length or low-latency streaming, where outputs must be emitted incrementally. Several works combine diffusion with autoregressive factorization for incremental generation~\cite{rasul2021autoregressive, wu2023ar, kim2024fifo, ruhe2024rolling}, but naive AR execution can reintroduce exposure bias and compounding errors through history. Diffusion Forcing~\cite{chen2024diffusion} addresses this by training with token-wise heterogeneous noise, improving robustness to imperfect histories. Follow-ups refine conditioning management: History-Guided Diffusion~\cite{songhistory} stabilizes coherence with corrupted history, Self Forcing~\cite{huang2025self} reduces train--test mismatch via self-generated history, and Rolling Forcing~\cite{liu2025rolling} targets real-time generation with rolling-window denoising. Nevertheless, most streaming diffusion systems assume static or slowly varying global conditions, and do not directly address interactive control with non-stationary signals such as streaming music, which requires balancing continuity and responsiveness.
% \section{Prelimiaries}
% \paragraph{Diffusion Models.}

% \paragraph{Vectorized Timestep Schedule.}
\section{Methods}
\label{sec:method}
As shown in Fig.~\ref{fig:pipeline}, our method is organized around three tightly coupled components for deployment-faithful streaming control: (i) \textbf{Causal Music Processing} that converts live audio into synchronized, strictly causal conditioning features; (ii) a \textbf{Streaming Audio-driven Motion Diffusion} model that performs bounded-latency, autoregressive diffusion generation over motion tokens/frames while remaining robust to imperfect histories; and (iii) a \textbf{Real-time Interactive System} that connects audio ingestion, online sampling, and avatar/robot playback in a closed timing loop with explicit latency budgets.

\subsection{Causal Music Processing}
\label{sec:causal_music}
\paragraph{Decomposed Music Conditioning.}
To support strictly streaming control without lookahead, we introduce a causal music encoding pipeline that extracts a compact feature $c_t$ from a fixed-length sliding audio buffer at each step $t$. This enables the motion generator to respond immediately to newly arriving musical cues and user edits. Concretely, we adopt a Vector-Quantized Periodic Autoencoder (VQ-PAE)~\cite{starke2022deepphase, li2024walkthedog} to factorize music conditioning into discrete rhythmic tokens and continuous phase-alignment dynamics, producing an embedding that is simultaneously beat-aware and temporally smooth—well-suited for low-latency, online motion synthesis.
At streaming step $t$, our generator is conditioned on a causal music feature
\begin{equation}
\mathbf{c}_t = [\mathbf{f}^{\text{vq}}_t; \ \mathbf{f}^{\text{pae}}_t],
\end{equation}
which is extracted from a fixed-length sliding audio window. The feature combines
(i) a discrete rhythmic code via residual vector quantization and (ii) a continuous periodic
alignment feature parameterized in the frequency domain.

\noindent\textbf{Discrete Rhythmic Encoding.}
Let $\mathbf{w}_t$ be the waveform segment in the sliding buffer at step $t$.
We compute a shared causal latent using a dilated 1D convolutional encoder:
\begin{equation}
\mathbf{f}^{\text{causal}} = \mathrm{Conv1D}(\mathbf{w}_t),
\end{equation}
where the dilation schedule is chosen to capture multi-scale temporal patterns while
preserving causality.
Then we feed the shared latent into a residual vector quantizer (RVQ)~\cite{van2017neural, lee2022autoregressive}:
\begin{equation}
\mathbf{f}^{\text{vq}} = \mathrm{RVQ}\!\left(\mathrm{FFN}_{\text{vq}}(\mathbf{f}^{\text{causal}})\right).
\end{equation}
This branch captures discrete rhythmic patterns and high-level musical structure that are
useful for motion triggering and style consistency.

\noindent\textbf{Periodic Alignment Encoding.}
To obtain a smooth temporal alignment signal, we estimate frequency-domain parameters
and predict the phase by a lightweight network using Fast Fourier Transformation (FFT)~\cite{starke2022deepphase, li2024walkthedog}:
\begin{equation}
[\mathbf{A},\mathbf{B},\mathbf{F}] = \mathrm{FFT}(\mathbf{f}^{\text{causal}}), \
\boldsymbol{\phi} = \tan^{-1}\left(\mathrm{FC}_{\text{phase}}(\mathbf{f}^{\text{causal}})\right).
\end{equation}
We reconstruct a periodic alignment feature in the time domain as:
\begin{equation}
\mathbf{f}^{\text{pae}}(t) =
\mathbf{A}\cdot \sin\!\big(2\pi(\mathbf{F}\cdot t-\boldsymbol{\phi})\big)+\mathbf{B},
\end{equation}
where $t$ denotes time within the window. Intuitively, $\boldsymbol{\phi}$ controls the alignment
(phase), while $\mathbf{A}$ and $\mathbf{B}$ control amplitude and offset for smooth transitions.

\subsection{Streaming Audio-driven Motion Diffusion}
We formulate real-time, audio-responsive interactive character control as a \emph{streaming} motion generation problem.
At the current time step $t$, our goal is to model the conditional distribution over the next $n$ motion frames,
$\mathbf{m}_{t:t+n-1}$, given a historical motion buffer $\mathbf{m}_{t-h:t-1}$ and the synchronized music context $\mathbf{c}_t$.
 
\paragraph{Streaming Motion Representation.}
Prior dance generation methods~\cite{tseng2023edge, li2024lodge} often model motion in global, non-canonical coordinates. In autoregressive (AR) rollouts, global-frame error accumulation shifts the conditioning distribution, degrading long-horizon stability (e.g., drift and jitter). We thus adopt a \emph{canonicalized incremental} representation that expresses dynamics in the root frame with velocity-like increments, yielding more stationary AR conditioning.

Many streaming generators~\cite{tevetclosd, chen2024taming} use the 263-D HumanML3D~\cite{guo2022generating} feature, which recovers joint positions but not the joint rotations needed for real-time retargeting, requiring costly post-processing~\cite{bogo2016keep}. To remove this bottleneck, we follow~\cite{xiao2025motionstreamer} and encode each frame as a 272-D vector:
\begin{equation}
\mathbf{m}_t=
\big\{
\dot r^{x}_t,\ \dot r^{z}_t,\ \dot r^{a}_t,\ \mathbf{j}^{p}_t,\ \mathbf{j}^{v}_t,\ \mathbf{j}^{r}_t
\big\}\in\mathbb{R}^{272},
\end{equation}
where $\dot r^{x}_t,\dot r^{z}_t\in\mathbb{R}$ are root XZ-plane linear velocities, $\dot r^{a}_t\in\mathbb{R}^{6}$ is the 6D root angular velocity~\cite{zhou2019continuity}, and $\mathbf{j}^{p}_t\in\mathbb{R}^{3K}$, $\mathbf{j}^{v}_t\in\mathbb{R}^{3K}$, $\mathbf{j}^{r}_t\in\mathbb{R}^{6K}$ are local joint positions, velocities, and rotations in the root frame. For SMPL~\cite{loper2015smpl}, $K=22$. This enables direct real-time reconstruction via forward kinematics, making AR streaming outputs immediately usable for low-latency retargeting without per-frame fitting.

\paragraph{Motion Primitive Learning.}
Given the strict low-latency requirements of streaming audio-driven motion generation, directly modeling high-dimensional raw motion dynamics is computationally prohibitive. To address this, we employ a motion Variational Autoencoder (VAE)~\cite{kingma2014auto} to project the motion stream into a compact latent manifold, aligned with prior frameworks~\cite{chen2023executing, zhaodartcontrol}. This strategy not only reduces the computational burden for the autoregressive backbone but also acts as an implicit low-pass filter, effectively removing high-frequency sensor jitter. Formally, let $\mathcal{T} = \{1, \dots, T\}$ denote the indices of the full motion sequence. The VAE consists of an encoder $\mathcal{E}$ and a decoder $\mathcal{D}$, which map the processed motion stream $\mathbf{m}_{\mathcal{T}} \in \mathbb{R}^{T \times 272}$ to a latent sequence $\mathbf{z}_{\mathcal{T}} \in \mathbb{R}^{T \times D_z}$ and back to an approximation $\hat{\mathbf{m}}_{\mathcal{T}}$, respectively:
\begin{equation}
    \mathbf{z}_{\mathcal{T}} = \mathcal{E}(\mathbf{m}_{\mathcal{T}}), \quad \hat{\mathbf{m}}_{\mathcal{T}} = \mathcal{D}(\mathbf{z}_{\mathcal{T}}).
\end{equation}

Our VAE is trained with an $\ell_2$ reconstruction loss to preserve motion fidelity and a Kullback-Leibler (KL) divergence loss to regularize the latent space towards a standard Gaussian prior:
\begin{equation}
    \begin{aligned}
        \mathcal{L}_{\text{VAE}} &= \|\mathbf{m}_{\mathcal{T}} -\hat{\mathbf{m}}_{\mathcal{T}}\|^2_2 \\ 
        &+ \lambda D_{\text{KL}}\big(q(\mathbf{z}_{\mathcal{T}}|\mathbf{m}_{\mathcal{T}}) \parallel \mathcal{N}(\mathbf{0}, \mathbf{I})\big),
    \end{aligned}
\end{equation}

where $\lambda$ is a weighting coefficient. 
\paragraph{Latent Diffusion Forcing.}
To enable streaming generation with flexible temporal control, we adapt the diffusion forcing framework~\cite{chen2024diffusion} to our latent sequence modeling.
We define a probability path specific to each temporal sequence index $t$ that transitions from clean data $\mathbf{x}_{t}^0 \equiv \mathbf{z}_{t}\sim q_{data}$ to pure noise $\mathbf{x}_t^{1} \sim \mathcal{N}(\mathbf{0}, \mathbf{I})$ over a diffusion horizon $k_t \in [0, 1]$:
\begin{equation}
    p(\mathbf{x}^{k_t}_{t} \mid \mathbf{x}^0_t) = \mathcal{N}(\mathbf{x}^{k_t}_{t}; \alpha_{k_t} \mathbf{x}^0_t, \sigma^2_{k_t} \mathbf{I}),
    \label{eq:forward}
\end{equation}
where $\alpha_{k_t}$ and $\sigma_{k_t}$ are predefined noise schedules satisfying the boundary conditions: $\alpha_0 = 1, \sigma_0=0$ and $\alpha_1 = 0, \sigma_1=1$.
This formulation allows each token at sequence position $t$ to undergo denoising at an independent noise level determined by $k_t$.
To reverse this process, based on the flow-based paradigm \cite{lipmanflow, li2025back}, we train a network $\mathbf{v}_\theta(\mathbf{x}^{k_{\mathcal{T}}}_{\mathcal{T}}, k_{\mathcal{T}}, \mathbf{c})$ to predict the conditional vector field $\mathbf{v}_{\mathcal{T}}$.
Each component $\mathbf{v}_t$ of this field corresponds to the instantaneous velocity of the probability flow with respect to the diffusion time:
\begin{equation} 
    \mathbf{v}_t = \dot{\alpha}_{k_t} \mathbf{x}^0_t + \dot{\sigma}_{k_t} \boldsymbol{\epsilon}_{t},
\end{equation}
where $\dot{\alpha}_{k_t}$ and $\dot{\sigma}_{k_t}$ denote the derivatives of the schedule coefficients with respect to $k_t$, and $\boldsymbol{\epsilon}_t \sim \mathcal{N}(\mathbf{0}, \mathbf{I})$.
The training objective minimizes the flow matching loss, computed selectively on noised tokens:
\begin{equation}
    \mathcal{L}_{\text{DF}} = \underset{{k_{\mathcal{T}}, \mathbf{z}_{\mathcal{T}}, \boldsymbol{\epsilon}_{\mathcal{T}}}}{\mathbb{E}} \left[ \|\mathbf{v}_\theta(\mathbf{x}^{k_{\mathcal{T}}}_{\mathcal{T}}, k_{\mathcal{T}}, \mathbf{c}) - \mathbf{v}_{\mathcal{T}} \|^2_{\mathcal{K}}\ \right],
\end{equation}
where the norm $\|\cdot\|_{\mathcal{K}}$ (defined as $\|\mathbf{u}\|^2_{\mathcal{K}} = \sum_{t: k_t > 0} \|u_t\|^2$ ) serves as a masking mechanism.
Since tokens with $k_t=0$ remain in the clean state and do not require denoising, this mask ensures the model focuses exclusively on learning the active transport dynamics.

\paragraph{Hybrid Temporal Schedule.}
Distinct from vanilla Diffusion Forcing which relies solely on random noise schedules, we employ a hybrid strategy to bridge the gap between training and streaming inference, while also encouraging consistent denoising patterns. At each iteration, we stochastically sample a noise schedule type from a categorical distribution defined over three variants:

1) \textbf{Random Schedule}: $k_t^{\text{rand}} \sim \mathcal{U}(0, 1)$. This independent sampling ensures the model retains the flexibility to handle arbitrary noise states.

2) \textbf{Monotonic Schedule}.
Motivated by \cite{zhang2024tedi, liu2025rolling}, we define a schedule based on a randomly sampled current time step $\tau \in [0, T]$ and a window size $l$:
\begin{equation}
    k_t^{\text{mono}} = \text{clamp}\left(\frac{t - (\tau - l)}{l}, 0, 1\right).
    \label{eq: mono}
\end{equation}
Under this schedule, frames before $\tau-l$ are clean ($k_t=0$), frames after $\tau$ are pure noise ($k_t=1$), and the intermediate window transitions linearly. 

3) \textbf{Trapezoid Schedule}.
Following \cite{chen2024diffusion, songhistory}, we further introduce a trapezoidal noise profile that assigns higher noise levels to tokens significantly preceding the current time step $\tau$:
\begin{equation}
\begin{aligned}
    k_t^{\text{hist}} = &\text{clamp}\left(\frac{(\tau - l_{\text{ctx}} - l) - t}{l_{\text{hist}}}, 0, 1\right), \\
    &k_t^{\text{trap}} = \max(k_t^{\text{hist}}, k_t^{\text{mono}}).
\end{aligned}
\label{eq: trap}
\end{equation}
Here, $l_{\text{ctx}}$ denotes the length of the clean context window and $l_{\text{hist}}>0$ determines the rate at which noise levels increase for the distant past.

\paragraph{Streaming Inference with Temporal Guidance.}
We perform bounded-latency streaming generation by maintaining a fixed-length FIFO denoising window of size $l$ at the tail of the sequence~\cite{kim2024fifo}, and apply hybrid noise schedule as temporal guidance to denoise all tokens in the window jointly (Alg.~\ref{alg:inference}).
% Let $\hat{\mathbf{x}}_{\mathcal{T}}$ denote the current latent sequence at the current denoising iteration.
% At a streaming step indexed by the current right boundary $\tau$, we denote the denoising window as $\mathcal{W}_\tau = \{\tau-l+1,\dots,\tau\}$.
% The prefix $\hat{\mathbf{x}}_{1:\tau-l}$ is treated as clean and used as conditioning context.
% During inference, we apply the monotonic schedule with respect to $\tau$ to assign noise levels $\{k_t\}_{t\in\mathcal{W}_\tau}$ that increase from left to right within the window, and then denoise all tokens in $\mathcal{W}_\tau$ simultaneously using the predicted velocity $\mathbf{v}_\theta(\hat{\mathbf{x}}^{k_{\mathcal{T}}}_{\mathcal{T}}, k_{\mathcal{T}}, \mathbf{c})_{\mathcal{W}_\tau}$.
Once the leftmost token in the window reaches the clean state, we emit it as the next streaming output token, then shift the buffer by one position and append a new token initialized as pure Gaussian noise, continuing the streaming process.

Inspired by \cite{ho2021classifier, songhistory}, we employ temporal guidance to mitigate long-term motion drift. By corrupting the distant history with a trapezoid noise schedule, we implicitly filter out high-frequency priors that may contradict current control signals. The guided velocity is then computed as a weighted sum with scale $\omega$:
\begin{equation}
    \begin{aligned}
    \mathbf{v}^{\text{guided}}_{\mathcal{T}}
    &= \mathbf{v}_\theta(\hat{\mathbf{x}}^{k_{\mathcal{T}}^{\text{mono}}}_{\mathcal{T}}, {k_{\mathcal{T}}^{\text{mono}}}, \varnothing)\\  
    &+ \omega[\mathbf{v}_\theta(\hat{\mathbf{x}}^{k_{\mathcal{T}}^{\text{trap}}}_{\mathcal{T}}, k_{\mathcal{T}}^{\text{trap}}, \mathbf{c}) -\mathbf{v}_\theta(\hat{\mathbf{x}}^{k_{\mathcal{T}}^{\text{mono}}}_{\mathcal{T}}, {k_{\mathcal{T}}^{\text{mono}}}, \varnothing)].
    \end{aligned}
\end{equation}

\begin{algorithm}[t]
\caption{Streaming Inference with Temporal Guidance}
\label{alg:inference}
\begin{algorithmic}[0]
\STATE \textbf{Input:} Audio stream $\mathbf{c}$, Model $\mathbf{v}_\theta$, Max length $T$, Window size $l$, Guidance scale $\omega$, Schedule params $l_{\text{ctx}}, l_{\text{hist}}$.
\STATE \textbf{Initialize:} Latent buffer $\hat{\mathbf{x}} = \varnothing$, Global step $\tau \leftarrow 0$, Solver step $\delta \leftarrow 1 / l$.  

\textcolor{gray}{\COMMENT{ensures one denoising step per output token}}

\WHILE{streaming}
    \STATE $\tau \leftarrow \tau + 1$.
    \STATE $\hat{\mathbf{x}} \leftarrow [\hat{\mathbf{x}}, \boldsymbol{\epsilon}]$, where $\boldsymbol{\epsilon} \sim \mathcal{N}(\mathbf{0}, \mathbf{I})$.
    \STATE $\hat{\mathbf{x}} \leftarrow \hat{\mathbf{x}}[-T:]$. \textcolor{gray}{\COMMENT{maintain the buffer}}
    
    \STATE Define active window $\mathcal{W} \leftarrow \{\tau - l + 1, \dots, \tau\}$.
    \STATE Define history indices $\mathcal{H} \leftarrow \{\text{indices before } \mathcal{W}\}$.
    
    \STATE $k^{\text{mono}}_t \leftarrow \Phi_{\text{mono}}(t, \tau, l)$ for $t \in \mathcal{W}$.
    \STATE $k^{\text{trap}}_t \leftarrow \Phi_{\text{trap}}(t, \tau, l, l_{\text{ctx}}, l_{\text{hist}})$ for all $t$.
    
    \textcolor{gray}{\COMMENT{construct the schedule}}
    \STATE $\hat{\mathbf{x}}^{\text{trap}} \leftarrow \hat{\mathbf{x}}$; Sample $\boldsymbol{\epsilon}' \sim \mathcal{N}(\mathbf{0}, \mathbf{I})$.

    \STATE $\hat{\mathbf{x}}^{\text{trap}}_{\mathcal{H}} \leftarrow \alpha(k^{\text{trap}}_{\mathcal{H}}) \odot \hat{\mathbf{x}}_{\mathcal{H}} + \sigma(k^{\text{trap}}_{\mathcal{H}}) \odot \boldsymbol{\epsilon}'_{\mathcal{H}}$.  
    
    \textcolor{gray}{\COMMENT{here $\odot$ denotes element-wise multiplication}}
    \STATE $\mathbf{v}_{\text{cond}} \leftarrow \mathbf{v}_\theta(\hat{\mathbf{x}}^{\text{trap}}, k^{\text{trap}}, \mathbf{c})$.
    \STATE $\mathbf{v}_{\text{hist}} \leftarrow \mathbf{v}_\theta(\hat{\mathbf{x}}, k^{\text{mono}}, \varnothing)$.
    \STATE $\mathbf{v} \leftarrow \mathbf{v}_{\text{hist}} + \omega (\mathbf{v}_{\text{cond}} - \mathbf{v}_{\text{hist}})$. \textcolor{gray}{\COMMENT{temporal guidance}}

    \STATE $\hat{\mathbf{x}}_{\mathcal{W}} \leftarrow \hat{\mathbf{x}}_{\mathcal{W}} - \mathbf{v}_{\mathcal{W}} \cdot \delta$. \textcolor{gray}{\COMMENT{denoise the window}}

    \STATE Emit token $\hat{\mathbf{x}}_{\tau-l+1}$ as output $\hat{\mathbf{z}}$.
\ENDWHILE
\end{algorithmic}
\end{algorithm}
% \section{Applications}
% \label{sec:app}

\begin{figure*}[t]
    \centering
    \includegraphics[width=\linewidth]{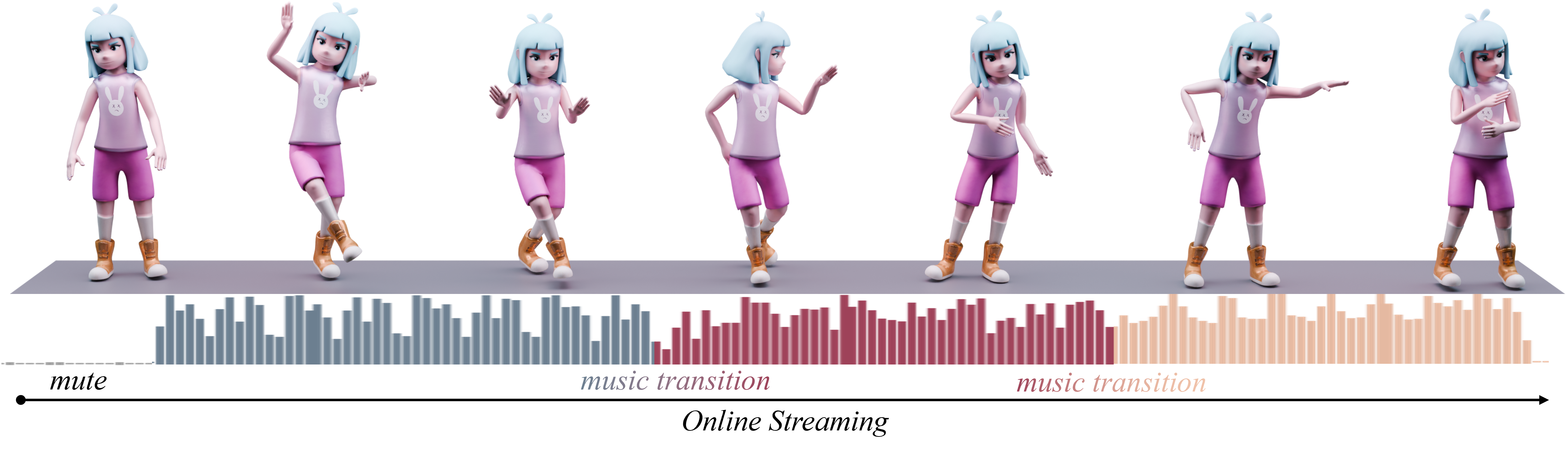}
    \caption{\textbf{Demonstration of Our Method.} In a strictly causal, bounded-latency online streaming rollout, DiscoForcing keeps the character stationary during silent segments (mute), and immediately generates beat-synchronized full-body dance once music resumes. As the input stream undergoes multiple music transitions, our model adapts in real time to the changing audio while maintaining long-horizon temporal coherence and smooth motion continuity.}
    \label{fig:demo}
\end{figure*}

\subsection{Real-Time Interactive System}
\label{sec:system}
To validate our streaming formulation beyond offline metrics, we build an end-to-end real-time interactive system that supports both virtual and physical deployment. As illustrated in Fig.~\ref{fig:pipeline}, the system exposes two application front-ends: (i) an \textbf{online avatar interaction platform} in Unity for user-facing animation interaction and (ii) a \textbf{physics-based humanoid deployment pipeline} targeting the Unitree G1 humanoid robot for embodied execution. The streaming generator communicates with both Unity and the G1 robot exclusively via ROS2.

\paragraph{Online Avatar Platform.}
% Our online avatar platform targets interactive animation and virtual visualization. It receives the decoded motion stream from the motion interface via ROS2, performs mesh- and skeleton-level retargeting to a target character rig, and renders the resulting pose in Unity for real-time user inspection. In practice, the platform maintains an animation-friendly representation by (i) mapping the predicted joint rotations to the avatar skeleton, (ii) retargeting to the skinned mesh while preserving kinematic constraints, and (iii) applying lightweight interpolation/resampling when the rendering loop runs at a different rate than the 30\,Hz motion stream. The Unity front-end further integrates an audio manager for synchronized playback and an on-screen motion manager for debugging and qualitative evaluation.
We implement an interactive animation and visualization platform in Unity, where virtual characters are controlled using SMPL parameters~\cite{loper2015smpl}. The streaming generator transmits real-time motion data to Unity through ROS2, enabling the characters to perform synchronized movements with the input music. This setup allows for live, user-facing evaluation of the generated motions under realistic streaming conditions.

\paragraph{Physics-Based Humanoid Platform.}
% We further deploy our streaming outputs on a real humanoid to demonstrate physical-world interaction. Starting from the predicted SMPL-space motion, we solve for the robot joint configuration using a GMR~\cite{araujo2025retargeting}-based mapping to obtain robot DoF poses, solved by temporally consistent IK and interpolation to match the control frequency of the hardware stack. The resulting reference trajectories are then tracked by a whole-body control (WBC) policy trained on large-scale motion datasets~\cite{liao2025beyondmimic}, which operates at 50\,Hz and produces target joint commands. Finally, a low-level PD controller running at 200\,Hz executes these commands on the humanoid hardware (Unitree G1), bridging streaming motion generation and real-time humanoid actuation.

We further validate the streaming generator on a physical humanoid robot. SMPL-format motions produced by the streaming generator are first retargeted to the Unitree G1 using GMR~\cite{araujo2025retargeting}. The resulting reference trajectories are subsequently tracked by a whole-body control (WBC) policy trained on large-scale motion datasets in simulation~\cite{liao2025beyondmimic, mahmood2019amass, punnakkal2021babel}. To ensure temporal alignment, the retargeted motion sequences are interpolated to match the tracker’s 50Hz control frequency. All of these steps are performed in real time, enabling the robot to dynamically adapt its dance movements in response to continuously changing music.

%%%%%%%%%%%%%%%%%%%%%%%%%%%%%%%

\begin{table}[t]
\centering
\renewcommand{\arraystretch}{1.2}
\setlength{\tabcolsep}{5pt}
\caption{Quantitative comparison on FineDance~\cite{li2023finedance} dataset. GT stands for ground truth, A higher or lower value is better for $\uparrow$ or $\downarrow$, and $\rightarrow$ means the value closer to ground truth is better.}
% \vspace{-0.05in}
\label{tab: comparison_finedance}
\scriptsize 
\resizebox{1.0\linewidth}{!}{
\begin{tabular}{l c c c  c c  c}
\toprule

 & FID$_{k}$$\downarrow$ & FID$_{g}$$\downarrow$ & FSR$\downarrow$
 & Div$_{k}$$\rightarrow$ & Div$_{g}$$\rightarrow$  & {BAS$\uparrow$} \\

\midrule
GT         
& --     & --     & 0.062
& 9.94 & 7.54 & 0.201  \\
\midrule

FACT
& 113.38 & 97.05 & 0.284
& 3.36  & \textbf{6.37} & 0.183 \\

Bailando 
& 82.81 & 28.17 & 0.188
& 7.74  & 6.25 & 0.202\\ 

EDGE      
& 94.34 & 50.38 & 0.200
& \textbf{8.13}  & 6.45 & 0.212 \\ 

Lodge   
& 50.00 &  35.52 & \textbf{0.028}
& 5.67  &  4.96 &  \textbf{0.226}\\ 

MEGA
& 50.00 & 13.02 & 0.243
& 6.23 & 6.27 & \textbf{0.226} \\

\midrule
\rowcolor{gray!20}
\textbf{DiscoForcing}
& \textbf{23.84} & \textbf{8.62} & 0.142
& 5.98 & 5.99 & 0.225\\

\bottomrule
\end{tabular}
}
\end{table}

\begin{table}[t]
\centering
\renewcommand{\arraystretch}{1.2}
\setlength{\tabcolsep}{5pt}
\caption{Comparison on AIST++~\cite{li2021ai} dataset.}
% \vspace{-0.05in}
\label{tab: comparison_aist}
\resizebox{1.0\linewidth}{!}{
\begin{tabular}{l  c c  c c  c}
\toprule
 & FID$_{k}$$\downarrow$ & FID$_{g}$$\downarrow$
 & Div$_{k}$$\rightarrow$ & Div$_{g}$$\rightarrow$
 & BAS$\uparrow$ \\

\midrule
Ground Truth          
& - & -
& 8.19 & 7.45
& 0.237 \\
\midrule

FACT
& 35.35 & 22.11
& 5.94 & 6.18
& 0.221 \\

Bailando
& 28.16 & \textbf{9.62}
& \textbf{7.83} & \textbf{6.34}
& 0.233 \\

EDGE   
& 42.16 & 22.12
& 3.96 & 4.61
& 0.233 \\

Lodge
& 37.09 & 18.79
& 5.58 & 4.85
& 0.242 \\

MEGA
& 25.89 & 12.62
& 5.84 & 6.23
& 0.238 \\

\midrule
\rowcolor{gray!20}
\textbf{DiscoForcing}
& \textbf{18.87} & 11.57
& 6.76 & 6.31
& \textbf{0.244} \\

\bottomrule
\end{tabular}
}
\end{table}

\begin{table*}[htbp]
    \caption[caption]{Ablation studies of music-to-dance generation on AIST++~\cite{li2021ai} dataset, where \underline{underlined} denotes the design we choose.}
    \begin{center}
    %\resizebox{1.0\textwidth}{!}{
        \begin{tabular}{l l  c c c  c c c  c}
        \toprule
        \multirow{2}{*}{Class} & \multirow{2}{*}{Settings}  & \multicolumn{3}{c}{Motion Quality} & \multicolumn{2}{c}{Motion Diversity} & \multirow{2}{*}{BAS$\uparrow$} & {Latency} \\
        \cmidrule(rl){3-5} \cmidrule(rl){6-7} 
         & & FID$_{k}$$\downarrow$ & FID$_{g}$$\downarrow$ & FSR$\downarrow$
         & Div$_{k}$$\rightarrow$ & Div$_{g}$$\rightarrow$  & &(ms / frame) \\
        \midrule
        \multicolumn{2}{c}{Ground Truth}  & --  & --  & 0.007  & 8.19  & 7.45  & 0.237  & --\\
        \midrule
        \multirow{3}{*}{Representation}
        & 151d & 27.83 & 15.78 & 0.201 & 5.57 & 6.62 & 0.233 & 25.65 \\
        & 263d & 26.47 & 8.30 & 0.060 & 8.53 & 7.62 & 0.245 & 26.60 \\
        & \underline{272d}  & 25.49 & 13.30 & 0.115 & 10.37 & 7.66 & 0.247 & 26.68 \\
        \midrule
        \multirow{2}{*}{Music Encoder} 
        & Librosa & 25.49 & 13.30 & 0.115 & 10.37 & 7.66 & 0.247 & 26.68 \\
        & \underline{VQ-PAE}  & 23.23 & 12.28 & 0.097 & 6.50 & 6.68 & 0.238 & 26.73 \\

        \midrule
        \multirow{2}{*}{Guidance} 
        & CFG & 23.23 & 12.28 & 0.097 & 6.50 & 6.68 & 0.238 & 26.73 \\
        & \underline{TG}  & 18.87 & 11.57 & 0.059 & 6.76 & 6.31 & 0.244 & 26.26 \\
        \midrule
        \multirow{3}{*}{Time-steps} 
        & 5   & 28.63 & 12.23 & 0.062 & 5.41 & 5.85 & 0.242 & 14.02\\
        & \underline{10}  & 18.87 & 11.57 & 0.059 & 6.76 & 6.31 & 0.244 & 26.26\\
        & 100 & 17.58 & 11.29 & 0.080 & 8.99 & 6.86 & 0.248 & 261.91 \\
        \midrule
        \rowcolor{gray!20}
        \multicolumn{2}{c}{DiscoForcing} 
          & 18.87 & 11.57 & 0.059 & 6.76 & 6.31 & 0.244 & 26.26 \\
        \bottomrule
        \end{tabular}
    %}
    \end{center}
    \label{tab:ablations}
\end{table*}

\section{Experiments}
\label{sec:exp}

\subsection{Experimental Setup}
\paragraph{Datasets.}
To validate our method, we utilize two public dance datasets that provide 3D dance motions synchronized with accompanying music: FineDance \cite{li2023finedance} and AIST++ \cite{li2021ai}. 
\textbf{FineDance} stands as the largest publicly available dataset to date, providing 7.7 hours of music-paired optical motion capture data at 30 FPS.
\textbf{AIST++} serves as a widely used standard benchmark, providing 5.2 hours of motion sequences reconstructed from multi-view videos at 60 FPS. 
For both datasets, we strictly follow the official data split for fair comparison and cut all examples to 30 FPS.

\paragraph{Evaluation Metrics.}
Following established protocols~\cite{yang2025megadance, li2024lodge, tseng2023edge}, we evaluate: \textbf{1) Motion Quality} (realism and physical plausibility), \textbf{2) Motion Diversity} (variety and non-repetitiveness), and \textbf{3) Music--Dance Correlation} (rhythmic synchronization).
For Motion Quality, we use Fréchet Inception Distance (FID)~\cite{heusel2017gans} to measure distributional similarity between generated and real motions. Following~\cite{li2021ai}, we report \textbf{FID$_k$} and \textbf{FID$_g$} on kinetic and geometric features, capturing physical smoothness and choreographic quality. We also adopt \textbf{FSR}~\cite{li2024lodge} to penalize foot-skating via consistency between center-of-mass acceleration and foot velocity. For Motion Diversity, we compute the average pairwise Euclidean distance of generated motion features and report \textbf{Div$_k$} and \textbf{Div$_g$} in the kinetic and geometric spaces~\cite{siyao2022bailando}. For Music--Dance Correlation, we use Beat Alignment Score (\textbf{BAS})~\cite{li2021ai} to quantify alignment between kinematic motion beats and musical audio beats.

\subsection{Music-to-Dance Evaluation}
\paragraph{Baselines.}
We compare our method against several representative music-to-dance approaches: FACT~\cite{li2021ai}, an autoregressive cross-modal transformer; Bailando~\cite{siyao2022bailando}, a VQ-VAE with actor--critic GPT framework; EDGE~\cite{tseng2023edge}, a diffusion transformer conditioned on Jukebox audio features; Lodge~\cite{li2024lodge}, a coarse-to-fine diffusion model with primitive generation and refinement; and MEGADance~\cite{yang2025megadance}, a Mixture-of-Experts Mamba--Transformer hybrid.

\paragraph{Comparisons.}
As summarized in Table~\ref{tab: comparison_finedance} and Table~\ref{tab: comparison_aist}, DiscoForcing achieves the best overall motion quality on FineDance, with substantially lower FID$_k$ and FID$_g$ than all baselines while keeping competitive BAS and diversity; on AIST++, DiscoForcing further improves kinetic FID$_k$ over strong diffusion and MoE baselines, indicating better rollout stability.
Notably, methods optimized for offline/global-context generation may achieve strong alignment or specific sub-metrics, but DiscoForcing provides a more balanced trade-off between long-horizon coherence and responsiveness in a deployment-faithful setting.

\subsection{Ablation Study}
\textbf{Motion Representation.}
As shown in Table~\ref{tab:ablations}, we compare our 272-dim representation with the 263-dim HumanML3D format and a 151-dim baseline. While 263-dim offers slight geometric gains, it lacks rotations and requires costly inverse kinematics, violating real-time constraints. Our 272-dim design best balances quality and deployability, improving fidelity and beat alignment over 151-dim while enabling low-latency forward-kinematics reconstruction.

\textbf{Music Encoder.}
Hand-crafted Librosa~\cite{mcfee2015librosa} features yield marginally better beat alignment due to explicit onset cues, but our learned VQ-PAE substantially improves realism. This indicates rule-based descriptors capture low-level pulses, whereas VQ-PAE encodes richer stylistic and semantic cues for natural control.

\textbf{Sampling Strategy.}
We ablate inference settings. Temporal Guidance (TG) outperforms Classifier-Free Guidance (CFG) by suppressing autoregressive drift and improving synchronization. For compute, 10 denoising steps are optimal: 100 steps brings diminishing gains while exceeding real-time limits, and 5 steps harms plausibility. We therefore adopt 10-step denoising to sustain stable 30 FPS throughput.

\section{Conclusion}
\label{sec:conclusion}
We present DiscoForcing, a novel streaming framework for audio-driven interactive character control that operates under strict causality and low-latency constraints. The core contribution is an autoregressive diffusion forcing generator that conditions on a short motion history and synchronized causal music features, enabling long-horizon, real-time motion synthesis without requiring future context or offline planning. We further describe a practical end-to-end system pipeline that supports both online avatar visualization and humanoid deployment via standard retargeting and tracking control. Experiments demonstrate that our approach improves motion quality and temporal consistency under streaming rollouts, while remaining responsive to time-varying audio conditions such as rhythm changes and abrupt transitions. We hope this work serves as a step toward deployment-faithful generative sequence modeling for interactive embodied applications, and we expect future work to explore stronger robustness under distribution shift, richer user control signals, stronger physical feasibility constraints, broader real-world testing, and tighter integration with downstream controllers for safer and more reliable deployment.

\newpage

\section*{Acknowledgements}
This work was supported by National Natural Science Foundation of China (62406195, 62303319),
Shanghai Local College Capacity Building Program (23010503100), HPC Platform of ShanghaiTech University, Core Facility Platform of Computer
Science and Communication of ShanghaiTech University, and MoE Key Laboratory of Intelligent
Perception and Human-Machine Collaboration (ShanghaiTech University), and Shanghai Engineering Research Center of Intelligent Vision and Imaging.

% \section*{Impact Statement}
% This paper presents a streaming, audio-conditioned motion generation method for interactive character control. The work may benefit animation, virtual reality, and embodied AI research by enabling lower-latency, responsive motion synthesis. Potential risks include misuse for deceptive synthetic media when combined with identity-linked avatars, bias inherited from training data, and safety concerns if outputs are executed on physical robots without appropriate constraints. We recommend responsible deployment practices, including conservative control and safety limits for real hardware, careful dataset documentation, and appropriate handling of audio and licensing.

% In the unusual situation where you want a paper to appear in the
% references without citing it in the main text, use \nocite

\bibliography{example_paper}

@inproceedings{tevethuman,
  title={Human Motion Diffusion Model},
  author={Tevet, Guy and Raab, Sigal and Gordon, Brian and Shafir, Yoni and Cohen-or, Daniel and Bermano, Amit Haim},
  booktitle={The Eleventh International Conference on Learning Representations},
  year={2023}
}

@inproceedings{chen2023executing,
  title={Executing your commands via motion diffusion in latent space},
  author={Chen, Xin and Jiang, Biao and Liu, Wen and Huang, Zilong and Fu, Bin and Chen, Tao and Yu, Gang},
  booktitle={Proceedings of the IEEE/CVF conference on computer vision and pattern recognition},
  pages={18000--18010},
  year={2023}
}

@inproceedings{zhang2023generating,
  title={Generating human motion from textual descriptions with discrete representations},
  author={Zhang, Jianrong and Zhang, Yangsong and Cun, Xiaodong and Zhang, Yong and Zhao, Hongwei and Lu, Hongtao and Shen, Xi and Shan, Ying},
  booktitle={Proceedings of the IEEE/CVF conference on computer vision and pattern recognition},
  pages={14730--14740},
  year={2023}
}

@inproceedings{wangscaling,
  title={Scaling Large Motion Models with Million-Level Human Motions},
  author={Wang, Ye and Zheng, Sipeng and Cao, Bin and Wei, Qianshan and Zeng, Weishuai and Jin, Qin and Lu, Zongqing},
  booktitle={Forty-second International Conference on Machine Learning},
  year={2025}
}

@article{starke2022deepphase,
  title={Deepphase: Periodic autoencoders for learning motion phase manifolds},
  author={Starke, Sebastian and Mason, Ian and Komura, Taku},
  journal={ACM Transactions on Graphics (ToG)},
  volume={41},
  number={4},
  pages={1--13},
  year={2022},
  publisher={ACM New York, NY, USA}
}

@article{holden2017phase,
  title={Phase-functioned neural networks for character control},
  author={Holden, Daniel and Komura, Taku and Saito, Jun},
  journal={ACM Transactions on Graphics (TOG)},
  volume={36},
  number={4},
  pages={1--13},
  year={2017},
  publisher={ACM New York, NY, USA}
}

@inproceedings{juravsky2022padl,
  title={Padl: Language-directed physics-based character control},
  author={Juravsky, Jordan and Guo, Yunrong and Fidler, Sanja and Peng, Xue Bin},
  booktitle={SIGGRAPH Asia 2022 Conference Papers},
  pages={1--9},
  year={2022}
}

@inproceedings{barquero2024seamless,
  title={Seamless human motion composition with blended positional encodings},
  author={Barquero, German and Escalera, Sergio and Palmero, Cristina},
  booktitle={Proceedings of the IEEE/CVF Conference on Computer Vision and Pattern Recognition},
  pages={457--469},
  year={2024}
}

@article{fan2025align,
  title={Align Your Rhythm: Generating Highly Aligned Dance Poses with Gating-Enhanced Rhythm-Aware Feature Representation},
  author={Fan, Congyi and Guan, Jian and Zhao, Xuanjia and Xu, Dongli and Lin, Youtian and Ye, Tong and Feng, Pengming and Pan, Haiwei},
  journal={arXiv preprint arXiv:2503.17340},
  year={2025}
}

@inproceedings{huang2024beat,
  title={Beat-it: Beat-synchronized multi-condition 3d dance generation},
  author={Huang, Zikai and Xu, Xuemiao and Xu, Cheng and Zhang, Huaidong and Zheng, Chenxi and Qin, Jing and He, Shengfeng},
  booktitle={European conference on computer vision},
  pages={273--290},
  year={2024},
  organization={Springer}
}

@article{van2017neural,
  title={Neural discrete representation learning},
  author={Van Den Oord, Aaron and Vinyals, Oriol and others},
  journal={Advances in neural information processing systems},
  volume={30},
  year={2017}
}

@inproceedings{lee2022autoregressive,
  title={Autoregressive image generation using residual quantization},
  author={Lee, Doyup and Kim, Chiheon and Kim, Saehoon and Cho, Minsu and Han, Wook-Shin},
  booktitle={Proceedings of the IEEE/CVF conference on computer vision and pattern recognition},
  pages={11523--11532},
  year={2022}
}

@inproceedings{li2024walkthedog,
  title={Walkthedog: Cross-morphology motion alignment via phase manifolds},
  author={Li, Peizhuo and Starke, Sebastian and Ye, Yuting and Sorkine-Hornung, Olga},
  booktitle={ACM SIGGRAPH 2024 Conference Papers},
  pages={1--10},
  year={2024}
}

@inproceedings{dai2024motionlcm,
  title={Motionlcm: Real-time controllable motion generation via latent consistency model},
  author={Dai, Wenxun and Chen, Ling-Hao and Wang, Jingbo and Liu, Jinpeng and Dai, Bo and Tang, Yansong},
  booktitle={European Conference on Computer Vision},
  pages={390--408},
  year={2024},
  organization={Springer}
}

@inproceedings{cui2024anyskill,
  title={Anyskill: Learning open-vocabulary physical skill for interactive agents},
  author={Cui, Jieming and Liu, Tengyu and Liu, Nian and Yang, Yaodong and Zhu, Yixin and Huang, Siyuan},
  booktitle={Proceedings of the IEEE/CVF conference on computer vision and pattern recognition},
  pages={852--862},
  year={2024}
}

@inproceedings{chen2024taming,
  title={Taming diffusion probabilistic models for character control},
  author={Chen, Rui and Shi, Mingyi and Huang, Shaoli and Tan, Ping and Komura, Taku and Chen, Xuelin},
  booktitle={ACM SIGGRAPH 2024 Conference Papers},
  pages={1--10},
  year={2024}
}

@inproceedings{zhaodartcontrol,
  title={DartControl: A Diffusion-Based Autoregressive Motion Model for Real-Time Text-Driven Motion Control},
  author={Zhao, Kaifeng and Li, Gen and Tang, Siyu},
  booktitle={The Thirteenth International Conference on Learning Representations}, 
  year={2025}
}

@article{cai2025flooddiffusion,
  title={FloodDiffusion: Tailored Diffusion Forcing for Streaming Motion Generation},
  author={Cai, Yiyi and Wu, Yuhan and Li, Kunhang and Zhou, You and Zheng, Bo and Liu, Haiyang},
  journal={arXiv preprint arXiv:2512.03520},
  year={2025}
}

@article{araujo2025retargeting,
  title={Retargeting matters: General motion retargeting for humanoid motion tracking},
  author={Araujo, Joao Pedro and Ze, Yanjie and Xu, Pei and Wu, Jiajun and Liu, C Karen},
  journal={arXiv preprint arXiv:2510.02252},
  year={2025}
}

@article{liao2025beyondmimic,
  title={Beyondmimic: From motion tracking to versatile humanoid control via guided diffusion},
  author={Liao, Qiayuan and Truong, Takara E and Huang, Xiaoyu and Gao, Yuman and Tevet, Guy and Sreenath, Koushil and Liu, C Karen},
  journal={arXiv preprint arXiv:2508.08241},
  year={2025}
}

@inproceedings{xiao2025motionstreamer,
  title={Motionstreamer: Streaming motion generation via diffusion-based autoregressive model in causal latent space},
  author={Xiao, Lixing and Lu, Shunlin and Pi, Huaijin and Fan, Ke and Pan, Liang and Zhou, Yueer and Feng, Ziyong and Zhou, Xiaowei and Peng, Sida and Wang, Jingbo},
  booktitle={Proceedings of the IEEE/CVF International Conference on Computer Vision},
  pages={10086--10096},
  year={2025}
}

@article{loper2015smpl,
  title={SMPL: a skinned multi-person linear model},
  author={Loper, Matthew and Mahmood, Naureen and Romero, Javier and Pons-Moll, Gerard and Black, Michael J},
  journal={ACM Transactions on Graphics (TOG)},
  volume={34},
  number={6},
  pages={1--16},
  year={2015},
  publisher={ACM New York, NY, USA}
}

@inproceedings{zhou2019continuity,
  title={On the continuity of rotation representations in neural networks},
  author={Zhou, Yi and Barnes, Connelly and Lu, Jingwan and Yang, Jimei and Li, Hao},
  booktitle={Proceedings of the IEEE/CVF conference on computer vision and pattern recognition},
  pages={5745--5753},
  year={2019}
}

@inproceedings{tevetclosd,
  title={CLoSD: Closing the Loop between Simulation and Diffusion for multi-task character control},
  author={Tevet, Guy and Raab, Sigal and Cohan, Setareh and Reda, Daniele and Luo, Zhengyi and Peng, Xue Bin and Bermano, Amit Haim and van de Panne, Michiel},
  booktitle={The Thirteenth International Conference on Learning Representations},
  year={2025}
}

@article{wu2025uniphys,
  title={UniPhys: Unified Planner and Controller with Diffusion for Flexible Physics-Based Character Control},
  author={Wu, Yan and Karunratanakul, Korrawe and Luo, Zhengyi and Tang, Siyu},
  journal={arXiv preprint arXiv:2504.12540},
  year={2025}
}

@article{huang2025diffuse,
  title={Diffuse-cloc: Guided diffusion for physics-based character look-ahead control},
  author={Huang, Xiaoyu and Truong, Takara and Zhang, Yunbo and Yu, Fangzhou and Sleiman, Jean Pierre and Hodgins, Jessica and Sreenath, Koushil and Farshidian, Farbod},
  journal={ACM Transactions on Graphics (TOG)},
  volume={44},
  number={4},
  pages={1--12},
  year={2025},
  publisher={ACM New York, NY, USA}
}

@article{jiang2023motiongpt,
  title={Motiongpt: Human motion as a foreign language},
  author={Jiang, Biao and Chen, Xin and Liu, Wen and Yu, Jingyi and Yu, Gang and Chen, Tao},
  journal={Advances in Neural Information Processing Systems},
  volume={36},
  pages={20067--20079},
  year={2023}
}

@inproceedings{li2021ai,
  title={Ai choreographer: Music conditioned 3d dance generation with aist++},
  author={Li, Ruilong and Yang, Shan and Ross, David A and Kanazawa, Angjoo},
  booktitle={Proceedings of the IEEE/CVF international conference on computer vision},
  pages={13401--13412},
  year={2021}
}

@inproceedings{siyao2022bailando,
  title={Bailando: 3d dance generation by actor-critic gpt with choreographic memory},
  author={Siyao, Li and Yu, Weijiang and Gu, Tianpei and Lin, Chunze and Wang, Quan and Qian, Chen and Loy, Chen Change and Liu, Ziwei},
  booktitle={Proceedings of the IEEE/CVF Conference on Computer Vision and Pattern Recognition},
  pages={11050--11059},
  year={2022}
}

@inproceedings{tseng2023edge,
  title={Edge: Editable dance generation from music},
  author={Tseng, Jonathan and Castellon, Rodrigo and Liu, Karen},
  booktitle={Proceedings of the IEEE/CVF conference on computer vision and pattern recognition},
  pages={448--458},
  year={2023}
}

@inproceedings{li2023finedance,
  title={Finedance: A fine-grained choreography dataset for 3d full body dance generation},
  author={Li, Ronghui and Zhao, Junfan and Zhang, Yachao and Su, Mingyang and Ren, Zeping and Zhang, Han and Tang, Yansong and Li, Xiu},
  booktitle={Proceedings of the IEEE/CVF International Conference on Computer Vision},
  pages={10234--10243},
  year={2023}
}

@inproceedings{li2024lodge,
  title={Lodge: A coarse to fine diffusion network for long dance generation guided by the characteristic dance primitives},
  author={Li, Ronghui and Zhang, YuXiang and Zhang, Yachao and Zhang, Hongwen and Guo, Jie and Zhang, Yan and Liu, Yebin and Li, Xiu},
  booktitle={Proceedings of the IEEE/CVF Conference on Computer Vision and Pattern Recognition},
  pages={1524--1534},
  year={2024}
}

@inproceedings{zhang2024bidirectional,
  title={Bidirectional autoregessive diffusion model for dance generation},
  author={Zhang, Canyu and Tang, Youbao and Zhang, Ning and Lin, Ruei-Sung and Han, Mei and Xiao, Jing and Wang, Song},
  booktitle={Proceedings of the IEEE/CVF Conference on Computer Vision and Pattern Recognition},
  pages={687--696},
  year={2024}
}

@inproceedings{yang2025megadance,
   title={{MEGAD}ance: Mixture-of-Experts Architecture for Genre-Aware 3D Dance Generation},
   author={Yang, Kaixing and Tang, Xulong and Peng, Ziqiao and Hu, Yuxuan and He, Jun and Liu, Hongyan},
   booktitle={The Thirty-ninth Annual Conference on Neural Information Processing Systems},
   year={2025},
}

@article{yang2025flowerdance,
  title={FlowerDance: MeanFlow for Efficient and Refined 3D Dance Generation},
  author={Yang, Kaixing and Tang, Xulong and Peng, Ziqiao and Zhang, Xiangyue and Wang, Puwei and He, Jun and Liu, Hongyan},
  journal={arXiv preprint arXiv:2511.21029},
  year={2025}
}

@inproceedings{guo2020action2motion,
  title={Action2motion: Conditioned generation of 3d human motions},
  author={Guo, Chuan and Zuo, Xinxin and Wang, Sen and Zou, Shihao and Sun, Qingyao and Deng, Annan and Gong, Minglun and Cheng, Li},
  booktitle={Proceedings of the 28th ACM international conference on multimedia},
  pages={2021--2029},
  year={2020}
}

@inproceedings{petrovich2021action,
  title={Action-conditioned 3d human motion synthesis with transformer vae},
  author={Petrovich, Mathis and Black, Michael J and Varol, G{\"u}l},
  booktitle={Proceedings of the IEEE/CVF international conference on computer vision},
  pages={10985--10995},
  year={2021}
}

@article{alexanderson2023listen,
  title={Listen, denoise, action! audio-driven motion synthesis with diffusion models},
  author={Alexanderson, Simon and Nagy, Rajmund and Beskow, Jonas and Henter, Gustav Eje},
  journal={ACM Transactions on Graphics (TOG)},
  volume={42},
  number={4},
  pages={1--20},
  year={2023},
  publisher={ACM New York, NY, USA}
}

@inproceedings{guo2022generating,
  title={Generating diverse and natural 3d human motions from text},
  author={Guo, Chuan and Zou, Shihao and Zuo, Xinxin and Wang, Sen and Ji, Wei and Li, Xingyu and Cheng, Li},
  booktitle={Proceedings of the IEEE/CVF conference on computer vision and pattern recognition},
  pages={5152--5161},
  year={2022}
}

@inproceedings{bogo2016keep,
  title={Keep it SMPL: Automatic estimation of 3D human pose and shape from a single image},
  author={Bogo, Federica and Kanazawa, Angjoo and Lassner, Christoph and Gehler, Peter and Romero, Javier and Black, Michael J},
  booktitle={European conference on computer vision},
  pages={561--578},
  year={2016},
  organization={Springer}
}

@inproceedings{sohl2015deep,
  title={Deep unsupervised learning using nonequilibrium thermodynamics},
  author={Sohl-Dickstein, Jascha and Weiss, Eric and Maheswaranathan, Niru and Ganguli, Surya},
  booktitle={International conference on machine learning},
  pages={2256--2265},
  year={2015},
  organization={pmlr}
}

@article{ho2020denoising,
  title={Denoising diffusion probabilistic models},
  author={Ho, Jonathan and Jain, Ajay and Abbeel, Pieter},
  journal={Advances in neural information processing systems},
  volume={33},
  pages={6840--6851},
  year={2020}
}

@inproceedings{gupta2024photorealistic,
  title={Photorealistic video generation with diffusion models},
  author={Gupta, Agrim and Yu, Lijun and Sohn, Kihyuk and Gu, Xiuye and Hahn, Meera and Li, Fei-Fei and Essa, Irfan and Jiang, Lu and Lezama, Jos{\'e}},
  booktitle={European Conference on Computer Vision},
  pages={393--411},
  year={2024},
  organization={Springer}
}

@inproceedings{janner2022planning,
  title={Planning with Diffusion for Flexible Behavior Synthesis},
  author={Janner, Michael and Du, Yilun and Tenenbaum, Joshua and Levine, Sergey},
  booktitle={International Conference on Machine Learning},
  pages={9902--9915},
  year={2022},
  organization={PMLR}
}

@article{ho2022video,
  title={Video diffusion models},
  author={Ho, Jonathan and Salimans, Tim and Gritsenko, Alexey and Chan, William and Norouzi, Mohammad and Fleet, David J},
  journal={Advances in neural information processing systems},
  volume={35},
  pages={8633--8646},
  year={2022}
}

@article{dhariwal2021diffusion,
  title={Diffusion models beat gans on image synthesis},
  author={Dhariwal, Prafulla and Nichol, Alexander},
  journal={Advances in neural information processing systems},
  volume={34},
  pages={8780--8794},
  year={2021}
}

@inproceedings{rombach2022high,
  title={High-resolution image synthesis with latent diffusion models},
  author={Rombach, Robin and Blattmann, Andreas and Lorenz, Dominik and Esser, Patrick and Ommer, Bj{\"o}rn},
  booktitle={Proceedings of the IEEE/CVF conference on computer vision and pattern recognition},
  pages={10684--10695},
  year={2022}
}

@article{saharia2022photorealistic,
  title={Photorealistic text-to-image diffusion models with deep language understanding},
  author={Saharia, Chitwan and Chan, William and Saxena, Saurabh and Li, Lala and Whang, Jay and Denton, Emily L and Ghasemipour, Kamyar and Gontijo Lopes, Raphael and Karagol Ayan, Burcu and Salimans, Tim and others},
  journal={Advances in neural information processing systems},
  volume={35},
  pages={36479--36494},
  year={2022}
}

@inproceedings{rasul2021autoregressive,
  title={Autoregressive denoising diffusion models for multivariate probabilistic time series forecasting},
  author={Rasul, Kashif and Seward, Calvin and Schuster, Ingmar and Vollgraf, Roland},
  booktitle={International conference on machine learning},
  pages={8857--8868},
  year={2021},
  organization={PMLR}
}

@inproceedings{song2023loss,
  title={Loss-guided diffusion models for plug-and-play controllable generation},
  author={Song, Jiaming and Zhang, Qinsheng and Yin, Hongxu and Mardani, Morteza and Liu, Ming-Yu and Kautz, Jan and Chen, Yongxin and Vahdat, Arash},
  booktitle={International Conference on Machine Learning},
  pages={32483--32498},
  year={2023},
  organization={PMLR}
}

@article{wu2023ar,
  title={Ar-diffusion: Auto-regressive diffusion model for text generation},
  author={Wu, Tong and Fan, Zhihao and Liu, Xiao and Zheng, Hai-Tao and Gong, Yeyun and Jiao, Jian and Li, Juntao and Guo, Jian and Duan, Nan and Chen, Weizhu and others},
  journal={Advances in Neural Information Processing Systems},
  volume={36},
  pages={39957--39974},
  year={2023}
}

@article{kim2024fifo,
  title={Fifo-diffusion: Generating infinite videos from text without training},
  author={Kim, Jihwan and Kang, Junoh and Choi, Jinyoung and Han, Bohyung},
  journal={Advances in Neural Information Processing Systems},
  volume={37},
  pages={89834--89868},
  year={2024}
}

@inproceedings{ruhe2024rolling,
   title={Rolling Diffusion Models},
   author={David Ruhe and Jonathan Heek and Tim Salimans and Emiel Hoogeboom},
   booktitle={Forty-first International Conference on Machine Learning},
   year={2024}
}

@article{chen2024diffusion,
  title={Diffusion forcing: Next-token prediction meets full-sequence diffusion},
  author={Chen, Boyuan and Mart{\'\i} Mons{\'o}, Diego and Du, Yilun and Simchowitz, Max and Tedrake, Russ and Sitzmann, Vincent},
  journal={Advances in Neural Information Processing Systems},
  volume={37},
  pages={24081--24125},
  year={2024}
}

@inproceedings{songhistory,
  title={History-Guided Video Diffusion},
  author={Song, Kiwhan and Chen, Boyuan and Simchowitz, Max and Du, Yilun and Tedrake, Russ and Sitzmann, Vincent},
  booktitle={Forty-second International Conference on Machine Learning},
  year={2025}
}

@inproceedings{huang2025self,
   title={Self Forcing: Bridging the Train-Test Gap in Autoregressive Video Diffusion},
   author={Xun Huang and Zhengqi Li and Guande He and Mingyuan Zhou and Eli Shechtman},
   booktitle={The Thirty-ninth Annual Conference on Neural Information Processing Systems},
   year={2025},
}

@article{liu2025rolling,
  title={Rolling forcing: Autoregressive long video diffusion in real time},
  author={Liu, Kunhao and Hu, Wenbo and Xu, Jiale and Shan, Ying and Lu, Shijian},
  journal={arXiv preprint arXiv:2509.25161},
  year={2025}
}

@inproceedings{punnakkal2021babel,
  title={BABEL: Bodies, action and behavior with english labels},
  author={Punnakkal, Abhinanda R and Chandrasekaran, Arjun and Athanasiou, Nikos and Quiros-Ramirez, Alejandra and Black, Michael J},
  booktitle={Proceedings of the IEEE/CVF conference on computer vision and pattern recognition},
  pages={722--731},
  year={2021}
}

@article{mcfee2015librosa,
  title={librosa: Audio and music signal analysis in python.},
  author={McFee, Brian and Raffel, Colin and Liang, Dawen and Ellis, Daniel PW and McVicar, Matt and Battenberg, Eric and Nieto, Oriol},
  journal={SciPy},
  volume={2015},
  pages={18--24},
  year={2015}
}

@article{heusel2017gans,
  title={Gans trained by a two time-scale update rule converge to a local nash equilibrium},
  author={Heusel, Martin and Ramsauer, Hubert and Unterthiner, Thomas and Nessler, Bernhard and Hochreiter, Sepp},
  journal={Advances in neural information processing systems},
  volume={30},
  year={2017}
}

@inproceedings{lipmanflow,
  title={Flow Matching for Generative Modeling},
  author={Lipman, Yaron and Chen, Ricky TQ and Ben-Hamu, Heli and Nickel, Maximilian and Le, Matthew},
  booktitle={The Eleventh International Conference on Learning Representations},
  year={2023}
}

@article{li2025back,
  title={Back to basics: Let denoising generative models denoise},
  author={Li, Tianhong and He, Kaiming},
  journal={arXiv preprint arXiv:2511.13720},
  year={2025}
}

@inproceedings{zhang2024tedi,
  title={Tedi: Temporally-entangled diffusion for long-term motion synthesis},
  author={Zhang, Zihan and Liu, Richard and Hanocka, Rana and Aberman, Kfir},
  booktitle={ACM SIGGRAPH 2024 Conference Papers},
  pages={1--11},
  year={2024}
}

@inproceedings{ho2021classifier,
  title={Classifier-Free Diffusion Guidance},
  author={Ho, Jonathan and Salimans, Tim},
  booktitle={NeurIPS 2021 Workshop on Deep Generative Models and Downstream Applications},
  year={2021}
}

@inproceedings{mahmood2019amass,
  title={AMASS: Archive of motion capture as surface shapes},
  author={Mahmood, Naureen and Ghorbani, Nima and Troje, Nikolaus F and Pons-Moll, Gerard and Black, Michael J},
  booktitle={Proceedings of the IEEE/CVF international conference on computer vision},
  pages={5442--5451},
  year={2019}
}

@article{kingma2014auto,
  title={Auto-Encoding Variational Bayes},
  author={Kingma, Diederik P and Welling, Max},
  journal={stat},
  volume={1050},
  pages={1},
  year={2014}
}

@inproceedings{ji2025towards,
  title={Towards immersive human-x interaction: A real-time framework for physically plausible motion synthesis},
  author={Ji, Kaiyang and Shi, Ye and Jin, Zichen and Chen, Kangyi and Xu, Lan and Ma, Yuexin and Yu, Jingyi and Wang, Jingya},
  booktitle={Proceedings of the IEEE/CVF International Conference on Computer Vision},
  pages={10173--10183},
  year={2025}
}

@inproceedings{tang2024unified,
  title={A unified diffusion framework for scene-aware human motion estimation from sparse signals},
  author={Tang, Jiangnan and Wang, Jingya and Ji, Kaiyang and Xu, Lan and Yu, Jingyi and Shi, Ye},
  booktitle={Proceedings of the IEEE/CVF Conference on Computer Vision and Pattern Recognition},
  pages={21251--21262},
  year={2024}
}

@inproceedings{deng2026humanobject,
title={Human-Object Interaction via Automatically Designed {VLM}-Guided Motion Policy},
author={Zekai Deng and Ye Shi and Kaiyang Ji and Lan Xu and Shaoli Huang and Jingya Wang},
booktitle={The Fourteenth International Conference on Learning Representations},
year={2026}
}

@article{liu2025commanding,
  title={Commanding Humanoid by Free-form Language: A Large Language Action Model with Unified Motion Vocabulary},
  author={Liu, Zhirui and Ji, Kaiyang and Yang, Ke and Yu, Jingyi and Shi, Ye and Wang, Jingya},
  journal={arXiv preprint arXiv:2511.22963},
  year={2025}
}
\bibliographystyle{icml2026}

\end{document}